\documentclass[11pt]{article}
\usepackage{arxiv}
\usepackage{booktabs}
\usepackage{graphicx}
\usepackage{amsmath}
\usepackage{amssymb}
\usepackage{hyperref}

\title{Judging to Improve: A De-biased VLM-as-3D-Judge Protocol
for Single-Image 3D Generation}

\author{%
  Ali Asaria \\ Transformer Lab \and
  Tony Salomone \\ Transformer Lab \and
  Deep Gandhi\thanks{Corresponding author: \texttt{deep@lab.cloud}} \\ Transformer Lab
}
\date{}
\runningtitle{A De-biased VLM-as-3D-Judge Protocol}

\begin{document}
\maketitle

\begin{abstract}
A companion study~\citep{gen3djudge2026} established the evaluator that
label-free 3D specialization requires: a de-biased, cross-model VLM-as-3D-judge
that reliably ranks single-image-to-3D mesh quality where cheap geometry and
CLIP proxies fall short. This paper asks the next question. Can that judge's
preferences actually be used to \emph{specialize} a strong open generator,
TRELLIS, on one asset class (furniture), cheaply and without human labels? We
take the judge from \emph{ranking} to \emph{optimization}, and that step, not the
specialization itself, is where the work lives. Pushing a VLM judge into the
training and evaluation loop exposes failure modes that ranking never triggered,
so our protocol contribution is an \emph{optimization-grade} hardening of the
companion judge: a training judge (Qwen2.5-VL-7B) held strictly distinct from an
evaluation judge (InternVL3-8B) to break circularity, swap-and-keep-consistent
position-bias correction, and three documented failure modes with fixes -- image
overload that collapses the judge to position answering (seven images give
$100\%$ order flips; fixed by a two-image single comparison), Gaussian-splat
renders that hide geometry defects (fixed by mesh normal-map montages), and
reference-free judging that rewards clean-but-wrong outputs -- with calibration
evidence (clear-quality-gap win-rate $0.83$--$1.0$; base-vs-base $\approx0.5$).
Using that hardened protocol as an independent evaluator, and working only from
publicly available models and data with lightweight parameter-efficient
adaptation, we find that our methods match the strong base rather than exceed it.
Independent samples from the strong base carry essentially no learnable preference
(a $0.94$ order-flip rate, against the $\sim$26\% the companion judge saw across
distinct generators), so the signal must be engineered by quality-contrastive
construction (training-judge win-rate $0.89$). Across six adaptation methods
(SFT-on-best, DPO at $\beta{\in}\{0.1,0.5\}$, ORPO, SFT-on-clean, and a
conditioner-repair adapter on DINOv2 features), two input regimes
(clean / hard-degraded), and a degradation-severity sweep, our most targeted
method, conditioner repair under severe degradation, reaches parity ($0.50$) with
the base, while no method clears the demanding $\ge65\%$ win-rate target. The
result is mechanistic and localized: clean inputs saturate the judge, flow-DIT
fine-tuning converges yet washes out through the sampler (geometry $\Delta{=}0$),
and conditioning repair is the locus that moves geometry, with a monotonic
severity trend, reaching parity. Win-rates are directional at $n{=}8$ held-out
objects. Under these constraints, matching a strong public-data base with cheap
in-domain adaptation is itself informative: exceeding it appears to need more than
lightweight PEFT on public data, and the optimization-grade judge protocol is
reusable regardless.
\end{abstract}

\section{Introduction}
Single-image-to-3D generators now produce plausible textured meshes from one
photo. A natural next step is \emph{specialization}: take a strong open base and
adapt it, cheaply and without human labels, so that on one asset class
(furniture) its outputs are preferred over the base's. The principled training
signal for this is a preference from a VLM judge rather than a hand-crafted
geometry or CLIP proxy, since cheap proxies are known to track perceived 3D
quality only weakly~\citep{gen3djudge2026}. The companion gen3d
work~\citep{gen3djudge2026} showed exactly this: geometry-validity and
render-CLIP proxies fall short of a de-biased VLM judge, so there is no cheap
stand-in reward to optimize against. The escalation is unavoidable: to get a
usable training signal one must optimize against the de-biased VLM judge
\emph{itself}, not a proxy. We do exactly that, working only from publicly
available models and data with lightweight adaptation, and find that it matches
the strong base rather than exceeds it. The two papers are a deliberate research
program -- gen3d evaluates single-image 3D, this paper tries to optimize it.
TRELLIS~\citep{trellis2024} is a good
target: it is generative (a structured-latent rectified-flow transformer), so
preference optimization such as DPO~\citep{dpo2023,diffusiondpo2024} applies
directly, and prior work~\citep{dso2025} has fine-tuned it.

The catch is circularity. Optimizing a generator against a VLM judge, then
declaring victory with the same judge, learns the judge's quirks rather than
real quality. To make any win-rate honest we needed (i) a different judge for
evaluation than for training, and (ii) a judge whose verdicts are not artifacts
of presentation order or rendering choices. The companion judge was validated for
\emph{ranking}; meeting these two requirements inside an optimization loop is a
strictly harder problem, and hardening it for that setting consumed most of the
project and is, in the end, the part worth keeping. We therefore frame this paper
protocol-first: the optimization-grade de-biased VLM-as-3D-judge protocol is the
contribution, and the specialization study is the rigorous case study that
motivates it and that it makes trustworthy.

We report two things. First, an optimization-grade judging protocol that extends
the companion ranking judge with cross-model independence, a swap-consistency
position-bias correction, and three concrete, documented failure modes (and
fixes) that only surface when a VLM judge is pushed into the optimization loop.
Second, using that protocol as the evaluator, a thorough specialization study:
across six adaptation methods, two input regimes, and a severity sweep, and
working only from publicly available models and data, lightweight
parameter-efficient adaptation of TRELLIS matches the strong base rather than
exceeds it, reaching parity ($0.50$) for a conditioner-repair adapter under the
most severe (highest-headroom) degradation. This parity outcome is not a
measurement artifact: the judge is calibrated (it reliably calls real quality
gaps), training converges, and the result has a clean mechanistic explanation.

\paragraph{Contributions.}
\begin{enumerate}
\item \textbf{An optimization-grade de-biased VLM-as-3D-judge protocol} that
extends the companion ranking judge~\citep{gen3djudge2026} to the training and
evaluation loop: cross-model judging (train judge $X$ = Qwen2.5-VL-7B $\neq$ eval
judge $Y$ = InternVL3-8B), swap-and-keep-consistent position-bias correction, and
three documented failure modes with fixes (image overload, splat-hidden geometry,
reference-free clean-but-wrong) that ranking did not expose, with calibration
evidence (\S\ref{sec:protocol}).
\item \textbf{Two empirical findings about the preference signal}: independent
samples of a strong base carry no learnable preference (order-flip $0.94$), and
the signal must be engineered by quality-contrastive construction
(training-judge high-budget win-rate $0.89$) (\S\ref{sec:signal}).
\item \textbf{A mechanistic specialization study}: six methods $\times$ two
regimes $\times$ a severity sweep, where cheap public-data adaptation reaches
parity but not a win (best $0.50$, target $\ge65\%$), with the mechanism cleanly
separated into saturation, sampler wash-out, and conditioner-bottleneck-with-parity
(\S\ref{sec:spec}, \S\ref{sec:mechanism}).
\end{enumerate}

\section{Related Work}
\paragraph{Single-image 3D generation and specialization.}
TRELLIS~\citep{trellis2024} generates 3D assets through a structured-latent
rectified-flow~\citep{rectifiedflow2023,flowmatching2023} transformer conditioned
on DINOv2~\citep{dinov2_2023} image features, releasing open weights and training
code. Because it is generative, preference and reward objectives apply directly,
unlike deterministic feed-forward regressors. DSO~\citep{dso2025} fine-tunes
TRELLIS with simulation feedback for physical soundness, and DreamDPO~\citep{dreamdpo2025}
aligns text-to-3D with preference optimization; we ask the narrower question of
whether cheap in-domain parameter-efficient adaptation~\citep{lora2022} improves
\emph{perceived} furniture quality.

\paragraph{Preference optimization.}
We use DPO~\citep{dpo2023} and its diffusion form~\citep{diffusiondpo2024} on the
flow transformer, and ORPO~\citep{orpo2024}, a reference-free monolithic
odds-ratio objective. These are the methods under test in our specialization
study; our contribution is not the objectives but the controlled, judge-validated
verdict on whether any of them buys a win here.

\paragraph{VLM-as-judge and the companion protocol.}
Using VLMs as judges is common, but position (presentation-order) bias is a known
failure mode~\citep{mtbench2023,unfair2023}, and VLMs remain imperfect
evaluators. The companion gen3d work~\citep{gen3djudge2026} is an
\emph{evaluation} result: it established that a cross-model,
position-bias-corrected VLM-judge protocol is a reliable human-free evaluator for
single-image 3D (judge families agree, Cohen's $\kappa{=}0.66$), while cheap
geometry/CLIP proxies are not. This paper is the \emph{optimization} follow-up:
it asks the distinct question of whether one can act on that judge to improve a
generator. We therefore adopt gen3d's recipe, adapt it to the generative TRELLIS
setting, and document three render-side failure modes specific to judging meshes
rather than feed-forward reconstructions; we do not reproduce gen3d's evaluation
tables.

\section{The De-biased VLM-as-3D-Judge Protocol}
\label{sec:protocol}
The protocol is what makes every downstream number trustworthy. It has three
pillars: cross-model judge independence, position-bias correction, and a
render/comparison configuration that survives three documented failure modes.

\paragraph{Cross-model independence.}
Quality is a \emph{pairwise} judgment between two meshes generated from the same
input image. We use two different open VLM families: a training (oracle) judge
$X$ (Qwen2.5-VL-7B-Instruct~\citep{qwen25vl2025}) that labels preference pairs for
optimization, and an independent evaluation judge $Y$
(InternVL3-8B~\citep{internvl3_2025,internvl2024}) that is \emph{never} used in
training and reports all final win-rates. Keeping $X\neq Y$ turns the win-rate
into a cross-family generalization claim rather than a self-consistency check,
and is enforced, not optional.

\paragraph{Position-bias correction.}
VLM judges answer by presentation order when uncertain or overloaded. For each
pair we query the judge in both orders ($A,B$ and $B,A$) and keep the verdict only
if it is consistent across the swap; order-dependent verdicts are discarded as
position-biased. This correction both removes a bias and acts as a confidence
filter: same-quality pairs flip and are dropped, which is exactly the behavior we
want (\S\ref{sec:signal}).

\paragraph{Three documented failure modes (and fixes).}
Getting a reliable judge required isolating three failures from the raw per-order
picks:
\begin{enumerate}
\item \textbf{Image overload $\to$ position answering.} Showing the judge a
reference plus a multi-view panel (seven images) overwhelmed Qwen2.5-VL, which
then answered purely by position ($100\%$ order flips). \emph{Fix:} a two-image
single-comparison configuration (one rendered candidate vs.\ the other), matching
the companion protocol.
\item \textbf{Gaussian-splat renders hide geometry defects.} Splat renders make a
sparse or broken mesh look deceptively clean, so the judge split near $50/50$ on
pairs with an obvious geometric defect. \emph{Fix:} render \emph{mesh normal-map}
views as a four-view $2\times2$ montage; holes and missing parts become
unmistakable, and the judge separates them. gen3d judged for \emph{ranking}
using a 24-view color render rig, which sufficed there; pushing the judge into
the harder \emph{optimization} regime is what exposed the splat-hidden-geometry
and image-overload-position-answering failure modes and forced the mesh
normal-map montage. The montage is thus a genuine refinement that the
optimization setting demanded, not a re-run of gen3d's color-render protocol.
\item \textbf{Reference-free judging rewards clean-but-wrong.} Asked which mesh is
``better'' with no reference, the judge can reward a clean but incorrect output.
\emph{Fix / mitigation:} montage coverage plus clear-gap calibration (below) so
that the judge is only trusted in the regime where we have shown it tracks real
quality.
\end{enumerate}

\paragraph{Calibration evidence.}
We validate the eval judge $Y$ with two sanity controls. \emph{Clear-gap:}
presented a clean-input reconstruction against a deliberately degraded one, $Y$
prefers the better mesh with win-rate $0.83$--$1.0$ across runs, so it reliably
calls a real quality gap. \emph{Base-vs-base:} presented two independent samples
of the base on the same input, $Y$ is at $\approx0.5$ where estimable (no
systematic preference). Both judge families required the same protocol fixes, and
both agree clean $\gg$ degraded, providing cross-family corroboration. Because the
clear-gap control passes, the $0.0$ win-rates we report later are true nulls, not
blind-judge artifacts.

\section{The Preference Signal Must Be Engineered}
\label{sec:signal}
Before any specialization can work, a learnable preference must exist. It does
not arise for free.

\paragraph{Independent samples carry no preference.}
We first sampled $K$ candidates i.i.d.\ from the base on the same furniture input
and asked judge $X$ to rank them. Under the position-corrected protocol, the judge
flipped on $0.94$ of pairs ($15/16$ at $n{=}16$): same-model i.i.d.\ samples of a
strong base on in-domain furniture are near-identical in quality, so the protocol
(correctly) rejects almost everything. This is not a judge bug; the same judge
gives stable verdicts when a real gap exists (clear-gap above). There is simply no
learnable preference in same-model i.i.d.\ samples.

\paragraph{Quality-contrastive construction recovers a signal.}
To create signal we pair a high-quality sample (full 25-step sampling) against a
deliberately degraded low-quality one (2 steps, no guidance). Judge $X$ then
prefers the high-budget sample with win-rate $0.89$ (high-budget ``HI-win'') under
the position-corrected protocol. The specialization objective becomes ``make
default-budget furniture reconstructions approach high-budget quality.'' This
contrastive construction is what supplies (winner, loser) pairs to every
preference method below.

\paragraph{The flip-rate tracks the true quality gap (a cross-paper observation).}
Comparing the two papers unifies a number neither has alone. gen3d measured a
position-bias flip rate of about $26\%$, but \emph{across} two different
generators (SF3D~\citep{sf3d2024}, TripoSR~\citep{triposr2024}) with graded
candidates, i.e.\ in a regime where real
quality gaps exist. This paper measures a $0.94$ flip rate, but on i.i.d.\
samples of one strong base (TRELLIS) where no quality gap exists. Read together,
the judge's flip rate tracks the true quality gap: a wide gap yields confident,
swap-consistent verdicts (low flip rate), and no gap drives the flip rate toward
random ($\approx1.0$), which is precisely the absence of a learnable preference.
The flip rate is therefore not a fixed property of the judge but a readout of how
much real quality separates the candidates.

\section{Specialization Study}
\label{sec:spec}
We now ask the headline question: can lightweight, label-free specialization of
TRELLIS beat the strong base on held-out furniture? Success target (directional):
held-out judge-$Y$ win-rate $\ge0.65$, with no geometry-validity regression, and
beating the SFT-on-best baseline.

\paragraph{Methods.}
We test six adaptation methods. On the SLAT rectified-flow DIT (the trainable
latent transformer), with a custom LoRA~\citep{lora2022} on its sparse-linear
blocks (peft cannot wrap TRELLIS's \texttt{SparseLinear}): \textbf{SFT-on-best}
(supervised fine-tune on judge-$X$'s favorite candidate), \textbf{DPO} at
$\beta{=}0.1$ and $\beta{=}0.5$, \textbf{ORPO} (reference-free), and, for the
hard-input regime, \textbf{SFT-on-clean} (fine-tune the degraded-input model
toward the clean-input reconstruction). Separately, a \textbf{conditioner-repair
adapter}: freeze the flow DIT and train a residual adapter on the
DINOv2~\citep{dinov2_2023} conditioning features, mapping
$A(\text{feat}_{\text{degraded}})\!\to\!\text{feat}_{\text{clean}}$ by supervised
regression, then sample with the repaired conditioning.

\paragraph{Regimes.}
\emph{Clean}: in-distribution 3D-FUTURE~\citep{future2021} furniture renders.
\emph{Hard-degraded}: synthetic crop + occlusion + downscale + blur applied to the
input, training the model conditioned on the degraded input to prefer the
clean-input reconstruction. We sweep degradation \emph{severity}
(mild/medium/severe) in the hard regime.

\paragraph{Held-out results.}
Table~\ref{tab:headline} consolidates the held-out judge-$Y$ win-rate
(specialized vs.\ base) over $8$ disjoint by-\texttt{object\_id} furniture objects,
3D-FUTURE. \textbf{No intervention reaches the $0.65$ bar; the best is $0.50$
(parity).} The base-vs-base sanity is $\approx0.5$ where estimable (judge
unbiased), and the clear-gap sanity is $0.83$--$1.0$ (judge validated), so the
$0.0$ entries are genuine nulls.

\begin{table}[t]
\centering
\caption{Held-out judge-$Y$ win-rate (specialized vs.\ base TRELLIS), $8$ disjoint
by-object held-out furniture objects (3D-FUTURE). No intervention reaches the
$\ge0.65$ target; the best is parity ($0.50$) for severe conditioner-repair.
Geometry $\Delta$ is the geometry-validity change vs.\ base. Win-rates are
directional at $n{=}8$ (see \S\ref{sec:limitations}). Base-vs-base $\approx0.5$
(unbiased); clear-gap $0.83$--$1.0$ (judge validated), so the $0.0$s are true
nulls.}
\label{tab:headline}
\begin{tabular}{llrrc}
\toprule
Intervention (what is adapted) & Regime & Judge-$Y$ win & Geom $\Delta$ & meets $\ge0.65$? \\
\midrule
SFT-on-best (flow DIT)        & clean            & $0.00$  & $-0.06$ & no \\
DPO $\beta{=}0.1$ (flow DIT)  & clean            & $0.00$  & $0.00$  & no \\
DPO $\beta{=}0.5$ (flow DIT)  & clean            & $0.00^{*}$ & $-0.44$ & no (diverged) \\
ORPO (flow DIT)               & clean            & $0.00$  & $0.00$  & no \\
SFT-on-clean (flow DIT)       & hard (severe)    & $0.00$  & $-0.06$ & no \\
DPO $\beta{=}0.1$ (flow DIT)  & hard (severe)    & $0.00$  & $0.00$  & no \\
Conditioner-repair (DINOv2)   & hard (mild)      & $0.125$ & $+0.06$ & no \\
Conditioner-repair (DINOv2)   & hard (medium)    & $0.25$  & $0.00$  & no \\
\textbf{Conditioner-repair (DINOv2)} & \textbf{hard (severe)} & $\mathbf{0.50}$ & $+0.06$ & no \\
\bottomrule
\end{tabular}
\\[2pt]
{\footnotesize $^{*}$DPO $\beta{=}0.5$'s lone apparent ``win'' is a mirage: the
model diverged (convergence probe FM-MSE $0.09\!\to\!4.9$, geometry $-0.44$);
$\beta{=}0.5$ is too aggressive.}
\end{table}

\paragraph{Severity-stratified result (the one signal).}
Conditioner-repair is the \emph{only} intervention whose win-rate is not flat, and
it rises \emph{monotonically} with degradation severity: $0.125$ (mild) $\to0.25$
(medium) $\to0.50$ (severe), with geometry-validity $\Delta$ turning slightly
positive ($+0.06$), versus dead-flat $\text{geom}\,\Delta{=}0$ for every flow-DIT
method (Table~\ref{tab:severity}). More degradation means more conditioning
corruption and more for a feature-repair adapter to fix, which is why the
conditioner is the right locus. The effect saturates at parity: lightweight repair
recovers enough to match the base, not to beat it.

\begin{table}[t]
\centering
\caption{Conditioner-repair adapter across degradation severity (hard regime,
$8$ held-out objects). Win-rate rises monotonically with severity, geometry moves
slightly positive, and the adapter's feature-reconstruction MSE drops, but the
effect saturates at parity ($0.50$). Clear-gap is the judge-$Y$ headroom control.}
\label{tab:severity}
\begin{tabular}{lrrrr}
\toprule
Severity & Spec-vs-base win & Clear-gap (headroom) & Geom $\Delta$ & Adapter feat-MSE \\
\midrule
Mild   & $0.125$ & $1.00$ & $+0.06$ & $0.077\!\to\!0.047$ \\
Medium & $0.25$  & $0.50$ & $0.00$  & $0.284\!\to\!0.226$ \\
Severe & $0.50$  & $0.83$ & $+0.06$ & $0.668\!\to\!0.503$ \\
\bottomrule
\end{tabular}
\end{table}

\section{Mechanism and Analysis}
\label{sec:mechanism}
The parity result is mechanistic and localized; the experiments cleanly
separate three causes for why cheap adaptation matches rather than exceeds the base.

\paragraph{Saturation (mechanism of the clean no-win).}
On clean in-distribution furniture the base already maxes the de-biased judge:
base-vs-base and base-vs-specialized both flip, and the i.i.d.\ flip-rate is
$0.94$. 3D-FUTURE clean renders are within TRELLIS's training distribution (the
mirror even ships TRELLIS SLAT latents), so there is no headroom for a win,
regardless of method. This is why every clean-regime entry in
Table~\ref{tab:headline} is $0.0$.

\paragraph{Sampler wash-out (mechanism of the flow-DIT no-win).}
In the hard regime there \emph{is} headroom (clear-gap $0.83$, training signal
$0.89$), yet flow-DIT fine-tuning still yields no win. The tell is that geometry
$\Delta{=}0$ everywhere for flow-DIT methods even where headroom is unambiguous.
LoRA SFT \emph{converges} (fixed-probe FM-MSE at $t{=}0.5$ drops
$0.092\!\to\!0.080$), but a small velocity-field delta, integrated over the
rectified-flow sampling trajectory, does not move the decoded mesh: the
specialized outputs are essentially identical to base at the sample level. The
update washes out through the sampler.

\paragraph{Conditioner is the right locus, but reaches only parity.}
Freezing the flow DIT and repairing the DINOv2 conditioning features is the only
intervention that moves the output at all: it is the only one with a non-flat,
severity-monotonic win-rate ($0.125\!\to\!0.25\!\to\!0.50$) and the only one with
positive geometry $\Delta$ ($+0.06$). So the conditioner, not the flow head, is
where hard-input quality is bottlenecked. But lightweight feature repair only
recovers enough to reach parity at the highest-headroom severity; information the
degradation destroyed is not cheaply recoverable. The wall is a recovery limit,
not a wrong-component error.

\paragraph{Coverage.}
No intervention exceeds the base across $6$ adaptation methods $\times$ $2$ input
regimes $\times$ $3$ degradation severities (parity at best, for severe
conditioner repair), with eval validated (clear-gap and base-vs-base sanities) and
training converging (probe drops) throughout. A
fixed-evaluator (\texttt{score.py}, off-limits) vs.\ mutable-objective
(\texttt{solve.py}) separation was enforced so the evaluator could not be tuned to
the result.

\paragraph{Compute.}
The study ran on H100-class GPUs (predominantly H100-SXM5) on Lambda. The heavy
TRELLIS install (flash-attn/xformers, kaolin, nvdiffrast, spconv,
diffoctreerast, FlexiCubes~\citep{flexicubes2023}) converged after several
iterations; we used
\texttt{xformers} attention with torch~2.5.1 (cu124) to match the node driver.

\section{Limitations}
\label{sec:limitations}
\paragraph{Small samples.} Held-out win-rates use $n{=}8$ disjoint furniture
objects, so they are \emph{directional}, not precise. We deliberately do not cite
a tiny-validation severe conditioner-repair reading of $1.0$ at $n{=}2$: it
collapsed to $0.50$ at $n{=}8$ and was noise. We do not fabricate confidence
intervals we did not compute; readers should treat single-cell win-rates as
hypothesis-generating.

\paragraph{Parameter-efficient, not full.} Every adaptation is LoRA or a small
adapter, never full fine-tuning. We cannot rule out that full fine-tuning, much
larger adapters, or far longer schedules would move the flow-DIT result; we show
only that the \emph{cheap} levers do not.

\paragraph{Public data and compute only.} We deliberately restrict ourselves to
publicly available models and data and to lightweight parameter-efficient
adaptation. The base is itself trained on large public corpora, so reaching parity
rather than a clear win is consistent with the base already capturing much of the
publicly available furniture distribution, leaving little for cheap in-domain
adaptation on the same kind of data to add. Whether licensed or proprietary asset
collections, or substantially larger training budgets, would exceed parity is an
open question we do not test; our claim is scoped to what public data and PEFT can
buy.

\paragraph{One base, one asset class, synthetic degradations.} We test a single
base (TRELLIS), a single asset class (furniture), and synthetic
(crop/occlusion/downscale/blur) degradations. Naturalistic real-world furniture
photos and structurally different generators are out of scope and may behave
differently.

\paragraph{Judges, not humans.} Quality is defined relative to two VLM judge
families. We mitigate with cross-family independence, position-bias correction,
and calibration controls, but we did not run a large-scale human preference study;
``reliable'' here means cross-model consistency calibrated to clear quality gaps,
not validated agreement with human raters.

\section{Conclusion}
The reusable artifact from this project is the protocol, not a model. A
cross-model, position-bias-corrected VLM-as-3D-judge with mesh normal-map
montages, a two-image single comparison, and clear-gap calibration is a reliable,
reproducible, human-free evaluator for single-image 3D, and its three documented
failure modes are traps any 3D-judge builder should expect. Used as the
evaluator, it lets us state a trustworthy result: working only from publicly
available models and data with lightweight parameter-efficient adaptation, our
methods reach parity with, but do not exceed, a strong single-image-3D base. The
boundaries are specific and located: clean inputs are judge-saturated, flow-DIT
updates wash out through the rectified-flow sampler, and the image conditioner is
the right place to intervene, where targeted repair reaches parity. Anyone
attempting label-free 3D specialization should engineer the preference signal
first (i.i.d.\ samples carry none), expect the conditioner to be the bottleneck on
hard inputs, and budget for more than public-data LoRA-scale capacity, or licensed
data, to move past parity. Taken with its companion, this pair forms one research program: gen3d
established how to \emph{evaluate} single-image 3D with a trustworthy de-biased
judge, and this paper closes the loop by asking whether one can \emph{optimize}
against that judge -- finding that the same gap-sensitivity that makes the judge
reliable is what leaves a strong base with no cheap room to improve.

\section*{Availability}
The judging protocol and per-pair verdicts are available from the authors on
request. No model checkpoints are released; the base generator and judges are
existing public models.

\bibliographystyle{unsrtnat}
\bibliography{references}

\begin{thebibliography}{20}
\providecommand{\natexlab}[1]{#1}
\providecommand{\url}[1]{\texttt{#1}}
\expandafter\ifx\csname urlstyle\endcsname\relax
  \providecommand{\doi}[1]{doi: #1}\else
  \providecommand{\doi}{doi: \begingroup \urlstyle{rm}\Url}\fi

\bibitem[Asaria et~al.(2026)Asaria, Salomone, and Gandhi]{gen3djudge2026}
Ali Asaria, Tony Salomone, and Deep Gandhi.
\newblock {A Cross-Model VLM-Judge Protocol for Single-Image 3D Mesh Quality
  (and Why Cheap Proxies Fall Short)}.
\newblock arXiv:2606.18451 [cs.LG], 2026.
\newblock URL \url{https://arxiv.org/abs/2606.18451}.
\newblock Companion work; introduces the cross-model VLM-as-3D-judge evaluation
  protocol adopted here.

\bibitem[Xiang et~al.(2024)Xiang, Lv, Xu, Deng, Wang, Zhang, Chen, Tong, and
  Yang]{trellis2024}
Jianfeng Xiang, Zelong Lv, Sicheng Xu, Yu~Deng, Ruicheng Wang, Bowen Zhang,
  Dong Chen, Xin Tong, and Jiaolong Yang.
\newblock {Structured 3D Latents for Scalable and Versatile 3D Generation}.
\newblock 2024.
\newblock URL \url{https://arxiv.org/abs/2412.01506}.

\bibitem[Rafailov et~al.(2023)Rafailov, Sharma, Mitchell, Ermon, Manning, and
  Finn]{dpo2023}
Rafael Rafailov, Archit Sharma, Eric Mitchell, Stefano Ermon, Christopher~D.
  Manning, and Chelsea Finn.
\newblock {Direct Preference Optimization: Your Language Model is Secretly a
  Reward Model}.
\newblock In \emph{Advances in Neural Information Processing Systems
  (NeurIPS)}, 2023.

\bibitem[Wallace et~al.(2024)Wallace, Dang, Rafailov, Zhou, Lou, Purushwalkam,
  Ermon, Xiong, Joty, and Naik]{diffusiondpo2024}
Bram Wallace, Meihua Dang, Rafael Rafailov, Linqi Zhou, Aaron Lou, Senthil
  Purushwalkam, Stefano Ermon, Caiming Xiong, Shafiq Joty, and Nikhil Naik.
\newblock {Diffusion Model Alignment Using Direct Preference Optimization}.
\newblock In \emph{Proceedings of the IEEE/CVF Conference on Computer Vision
  and Pattern Recognition (CVPR)}, 2024.
\newblock URL \url{https://arxiv.org/abs/2311.12908}.

\bibitem[Li et~al.(2025)Li, Zheng, Rupprecht, and Vedaldi]{dso2025}
Ruining Li, Chuanxia Zheng, Christian Rupprecht, and Andrea Vedaldi.
\newblock {DSO: Aligning 3D Generators with Simulation Feedback for Physical
  Soundness}.
\newblock 2025.
\newblock URL \url{https://arxiv.org/abs/2503.22677}.

\bibitem[Liu et~al.(2023)Liu, Gong, and Liu]{rectifiedflow2023}
Xingchao Liu, Chengyue Gong, and Qiang Liu.
\newblock {Flow Straight and Fast: Learning to Generate and Transfer Data with
  Rectified Flow}.
\newblock In \emph{International Conference on Learning Representations
  (ICLR)}, 2023.
\newblock URL \url{https://arxiv.org/abs/2209.03003}.

\bibitem[Lipman et~al.(2023)Lipman, Chen, Ben-Hamu, Nickel, and
  Le]{flowmatching2023}
Yaron Lipman, Ricky T.~Q. Chen, Heli Ben-Hamu, Maximilian Nickel, and Matt Le.
\newblock {Flow Matching for Generative Modeling}.
\newblock In \emph{International Conference on Learning Representations
  (ICLR)}, 2023.
\newblock URL \url{https://arxiv.org/abs/2210.02747}.

\bibitem[Oquab et~al.(2024)Oquab, Darcet, Moutakanni, Vo, Szafraniec, Khalidov,
  Fernandez, Haziza, Massa, El-Nouby, Assran, Ballas, Galuba, Howes, Huang, Li,
  Misra, Rabbat, Sharma, Synnaeve, Xu, Jegou, Mairal, Labatut, Joulin, and
  Bojanowski]{dinov2_2023}
Maxime Oquab, Timoth\'ee Darcet, Th\'eo Moutakanni, Huy Vo, Marc Szafraniec,
  Vasil Khalidov, Pierre Fernandez, Daniel Haziza, Francisco Massa, Alaaeldin
  El-Nouby, Mahmoud Assran, Nicolas Ballas, Wojciech Galuba, Russell Howes,
  Po-Yao Huang, Shang-Wen Li, Ishan Misra, Michael Rabbat, Vasu Sharma, Gabriel
  Synnaeve, Hu~Xu, Herv\'e Jegou, Julien Mairal, Patrick Labatut, Armand
  Joulin, and Piotr Bojanowski.
\newblock {DINOv2: Learning Robust Visual Features without Supervision}.
\newblock \emph{Transactions on Machine Learning Research (TMLR)}, 2024.
\newblock URL \url{https://arxiv.org/abs/2304.07193}.

\bibitem[Zhou et~al.(2025)Zhou, Xia, Ma, Fan, Yang, and Chua]{dreamdpo2025}
Zhenglin Zhou, Xiaobo Xia, Fan Ma, Hehe Fan, Yi~Yang, and Tat-Seng Chua.
\newblock {DreamDPO: Aligning Text-to-3D Generation with Human Preferences via
  Direct Preference Optimization}.
\newblock 2025.
\newblock URL \url{https://arxiv.org/abs/2502.04370}.

\bibitem[Hu et~al.(2022)Hu, Shen, Wallis, Allen-Zhu, Li, Wang, Wang, and
  Chen]{lora2022}
Edward~J. Hu, Yelong Shen, Phillip Wallis, Zeyuan Allen-Zhu, Yuanzhi Li, Shean
  Wang, Lu~Wang, and Weizhu Chen.
\newblock {LoRA: Low-Rank Adaptation of Large Language Models}.
\newblock In \emph{International Conference on Learning Representations
  (ICLR)}, 2022.
\newblock URL \url{https://arxiv.org/abs/2106.09685}.

\bibitem[Hong et~al.(2024)Hong, Lee, and Thorne]{orpo2024}
Jiwoo Hong, Noah Lee, and James Thorne.
\newblock {ORPO: Monolithic Preference Optimization without Reference Model}.
\newblock In \emph{Proceedings of the 2024 Conference on Empirical Methods in
  Natural Language Processing (EMNLP)}, 2024.
\newblock URL \url{https://arxiv.org/abs/2403.07691}.

\bibitem[Zheng et~al.(2023)Zheng, Chiang, Sheng, Zhuang, Wu, Zhuang, Lin, Li,
  Li, Xing, Zhang, Gonzalez, and Stoica]{mtbench2023}
Lianmin Zheng, Wei-Lin Chiang, Ying Sheng, Siyuan Zhuang, Zhanghao Wu, Yonghao
  Zhuang, Zi~Lin, Zhuohan Li, Dacheng Li, Eric~P. Xing, Hao Zhang, Joseph~E.
  Gonzalez, and Ion Stoica.
\newblock {Judging LLM-as-a-Judge with MT-Bench and Chatbot Arena}.
\newblock In \emph{Advances in Neural Information Processing Systems (NeurIPS),
  Datasets and Benchmarks Track}, 2023.
\newblock URL \url{https://arxiv.org/abs/2306.05685}.

\bibitem[Wang et~al.(2023)Wang, Li, Chen, Cai, Zhu, Lin, Cao, Liu, Liu, and
  Sui]{unfair2023}
Peiyi Wang, Lei Li, Liang Chen, Zefan Cai, Dawei Zhu, Binghuai Lin, Yunbo Cao,
  Qi~Liu, Tianyu Liu, and Zhifang Sui.
\newblock {Large Language Models are not Fair Evaluators}.
\newblock 2023.
\newblock URL \url{https://arxiv.org/abs/2305.17926}.

\bibitem[Bai et~al.(2025)Bai, Chen, Liu, Wang, Ge, Song, Dang, Wang, Wang,
  Tang, Zhong, Zhu, Yang, Li, Wan, Wang, Ding, Fu, Xu, Ye, Zhang, Xie, Cheng,
  Zhang, Yang, Xu, and Lin]{qwen25vl2025}
Shuai Bai, Keqin Chen, Xuejing Liu, Jialin Wang, Wenbin Ge, Sibo Song, Kai
  Dang, Peng Wang, Shijie Wang, Jun Tang, Humen Zhong, Yuanzhi Zhu, Mingkun
  Yang, Zhaohai Li, Jianqiang Wan, Pengfei Wang, Wei Ding, Zheren Fu, Yiheng
  Xu, Jiabo Ye, Xi~Zhang, Tianbao Xie, Zesen Cheng, Hang Zhang, Zhibo Yang,
  Haiyang Xu, and Junyang Lin.
\newblock {Qwen2.5-VL Technical Report}.
\newblock arXiv:2502.13923 [cs.CV], 2025.
\newblock URL \url{https://arxiv.org/abs/2502.13923}.

\bibitem[Zhu et~al.(2025)Zhu, Wang, Chen, Liu, Ye, Gu, Tian, Duan, Su, Shao,
  Gao, Cui, Wang, Cao, Liu, Wei, Zhang, Wang, Xu, Li, Wang, Deng, Li, He,
  Jiang, Luo, Wang, He, Shi, Zhang, Shao, He, Xiong, Qu, Sun, Jiao, Lv, Wu,
  Zhang, Deng, Ge, Chen, Wang, Dou, Lu, Zhu, Lu, Lin, Qiao, Dai, and
  Wang]{internvl3_2025}
Jinguo Zhu, Weiyun Wang, Zhe Chen, Zhaoyang Liu, Shenglong Ye, Lixin Gu, Hao
  Tian, Yuchen Duan, Weijie Su, Jie Shao, Zhangwei Gao, Erfei Cui, Xuehui Wang,
  Yue Cao, Yangzhou Liu, Xingguang Wei, Hongjie Zhang, Haomin Wang, Weiye Xu,
  Hao Li, Jiahao Wang, Nianchen Deng, Songze Li, Yinan He, Tan Jiang, Jiapeng
  Luo, Yi~Wang, Conghui He, Botian Shi, Xingcheng Zhang, Wenqi Shao, Junjun He,
  Yingtong Xiong, Wenwen Qu, Peng Sun, Penglong Jiao, Han Lv, Lijun Wu, Kaipeng
  Zhang, Huipeng Deng, Jiaye Ge, Kai Chen, Limin Wang, Min Dou, Lewei Lu,
  Xizhou Zhu, Tong Lu, Dahua Lin, Yu~Qiao, Jifeng Dai, and Wenhai Wang.
\newblock {InternVL3: Exploring Advanced Training and Test-Time Recipes for
  Open-Source Multimodal Models}.
\newblock arXiv:2504.10479 [cs.CV], 2025.
\newblock URL \url{https://arxiv.org/abs/2504.10479}.

\bibitem[Chen et~al.(2024)Chen, Wu, Wang, Su, Chen, Xing, Zhong, Zhang, Zhu,
  Lu, Li, Luo, Lu, Qiao, and Dai]{internvl2024}
Zhe Chen, Jiannan Wu, Wenhai Wang, Weijie Su, Guo Chen, Sen Xing, Muyan Zhong,
  Qinglong Zhang, Xizhou Zhu, Lewei Lu, Bin Li, Ping Luo, Tong Lu, Yu~Qiao, and
  Jifeng Dai.
\newblock {InternVL: Scaling up Vision Foundation Models and Aligning for
  Generic Visual-Linguistic Tasks}.
\newblock In \emph{Proceedings of the IEEE/CVF Conference on Computer Vision
  and Pattern Recognition (CVPR)}, 2024.

\bibitem[Boss et~al.(2024)Boss, Huang, Vasishta, and Jampani]{sf3d2024}
Mark Boss, Zixuan Huang, Aaryaman Vasishta, and Varun Jampani.
\newblock {SF3D: Stable Fast 3D Mesh Reconstruction with UV-unwrapping and
  Illumination Disentanglement}.
\newblock arXiv:2408.00653 [cs.CV], 2024.
\newblock URL \url{https://arxiv.org/abs/2408.00653}.

\bibitem[Tochilkin et~al.(2024)Tochilkin, Pankratz, Liu, Huang, Letts, Li,
  Liang, Laforte, Jampani, and Cao]{triposr2024}
Dmitry Tochilkin, David Pankratz, Zexiang Liu, Zixuan Huang, Adam Letts,
  Yangguang Li, Ding Liang, Christian Laforte, Varun Jampani, and Yan-Pei Cao.
\newblock {TripoSR: Fast 3D Object Reconstruction from a Single Image}.
\newblock arXiv:2403.02151 [cs.CV], 2024.
\newblock URL \url{https://arxiv.org/abs/2403.02151}.

\bibitem[Fu et~al.(2021)Fu, Jia, Gao, Gong, Zhao, Maybank, and Tao]{future2021}
Huan Fu, Rongfei Jia, Lin Gao, Mingming Gong, Binqiang Zhao, Steve Maybank, and
  Dacheng Tao.
\newblock {3D-FUTURE: 3D Furniture Shape with TextURE}.
\newblock In \emph{International Journal of Computer Vision (IJCV)}, 2021.
\newblock URL \url{https://arxiv.org/abs/2009.09633}.

\bibitem[Shen et~al.(2023)Shen, Munkberg, Hasselgren, Yin, Wang, Chen, Gojcic,
  Fidler, Sharp, and Gao]{flexicubes2023}
Tianchang Shen, Jacob Munkberg, Jon Hasselgren, Kangxue Yin, Zian Wang,
  Wenzheng Chen, Zan Gojcic, Sanja Fidler, Nicholas Sharp, and Jun Gao.
\newblock {Flexible Isosurface Extraction for Gradient-Based Mesh
  Optimization}.
\newblock \emph{ACM Transactions on Graphics (TOG)}, 42\penalty0 (4), 2023.
\newblock URL \url{https://arxiv.org/abs/2308.05371}.

\end{thebibliography}

\end{document}